\documentclass{article}

\usepackage{microtype}
\usepackage{graphicx}
\usepackage{subcaption}
\usepackage{booktabs} % for professional tables

\usepackage{enumitem}
\usepackage{amsmath}
\usepackage{amssymb}
\usepackage{stmaryrd}
\usepackage{dsfont}
\usepackage{multirow}
\usepackage{comment}

\usepackage{amsthm}
\newtheorem*{claim}{Claim}

\usepackage{hyperref}
\usepackage[accepted]{icml2018}
\usepackage[colorinlistoftodos]{todonotes}

\newcommand{\supp}{the Appendix}

\newcommand{\cutsectionup}{\vspace*{-0.08in}}%{\vspace*{-0.05in}}
\newcommand{\cutsectiondown}{\vspace*{-0.05in}}%{\vspace*{-0.05in}}

\newcommand{\cutsubsectionup}{\vspace*{-0.05in}}%{\vspace*{-0.05in}}
\newcommand{\cutsubsectiondown}{\vspace*{-0.05in}}%{\vspace*{-0.05in}}

\newcommand{\cutparagraphup}{\vspace*{-0.05in}}%{\vspace*{-0.05in}}
\icmltitlerunning{Self-Imitation Learning}

\begin{document}

\twocolumn[
%\icmltitle{Self-Imitation Learning}
%\icmltitle{Self-Imitation Learning: On the Importance of Exploitation}
%\icmltitle{Self-Imitation Learning: On the Importance of Exploitation in Reinforcement Learning}
%\icmltitle{On the Importance of Exploitation in Deep Reinforcement Learning}
%\icmltitle{On the Importance of Exploitation in Complex Environments}
%\icmltitle{Self-Imitation Learning: On the Importance of Exploitation}
\icmltitle{Self-Imitation Learning}
%\icmltitle{Deep Exploration via Self-Imitation Learning}

\icmlsetsymbol{equal}{*}

\begin{icmlauthorlist}
\icmlauthor{Junhyuk Oh}{equal,umich}
\icmlauthor{Yijie Guo}{equal,umich}
\icmlauthor{Satinder Singh}{umich}
\icmlauthor{Honglak Lee}{google,umich}
\end{icmlauthorlist}

\icmlaffiliation{umich}{University of Michigan}
\icmlaffiliation{google}{Google Brain}

\icmlcorrespondingauthor{Junhyuk Oh}{junhyuk@umich.edu}
\icmlcorrespondingauthor{Yijie Guo}{guoyijie@umich.edu}

\icmlkeywords{Reinforcement Learning}

\vskip 0.3in
]



\printAffiliationsAndNotice{\icmlEqualContribution} % otherwise use the standard text.

\begin{abstract}
This paper proposes \textit{Self-Imitation Learning} (SIL), a simple off-policy actor-critic algorithm that learns to reproduce the agent's past {\em good} decisions. This algorithm is designed to verify our hypothesis that exploiting past good experiences can indirectly drive deep exploration. Our empirical results show that SIL significantly improves advantage actor-critic (A2C) on several hard exploration Atari games and is competitive to the state-of-the-art count-based exploration methods. We also show that SIL improves proximal policy optimization (PPO) on MuJoCo tasks.
\end{abstract}

\section{Introduction} \label{sec:intro}
\cutsectiondown
The trade-off between exploration and exploitation is one of the fundamental challenges in reinforcement learning (RL). The agent needs to \textit{exploit} what it already knows in order to maximize reward. But, the agent also needs to \textit{explore} new behaviors in order to find a potentially better policy. The resulting performance of an RL agent emerges from this interaction between exploration and exploitation.

\begin{figure}
    \centering
    \begin{subfigure}{0.38\linewidth}
    	\centering
	    \includegraphics[width=\textwidth]{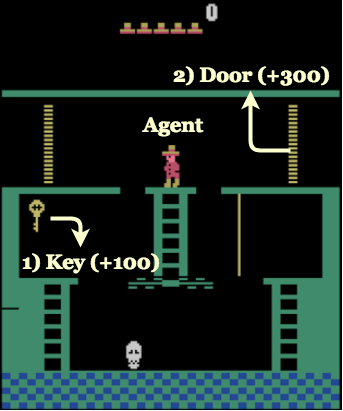} 
   	\end{subfigure}
   	\hspace{-7pt}
    \begin{subfigure}{0.6\linewidth}
    	\centering
	    \includegraphics[width=\textwidth]{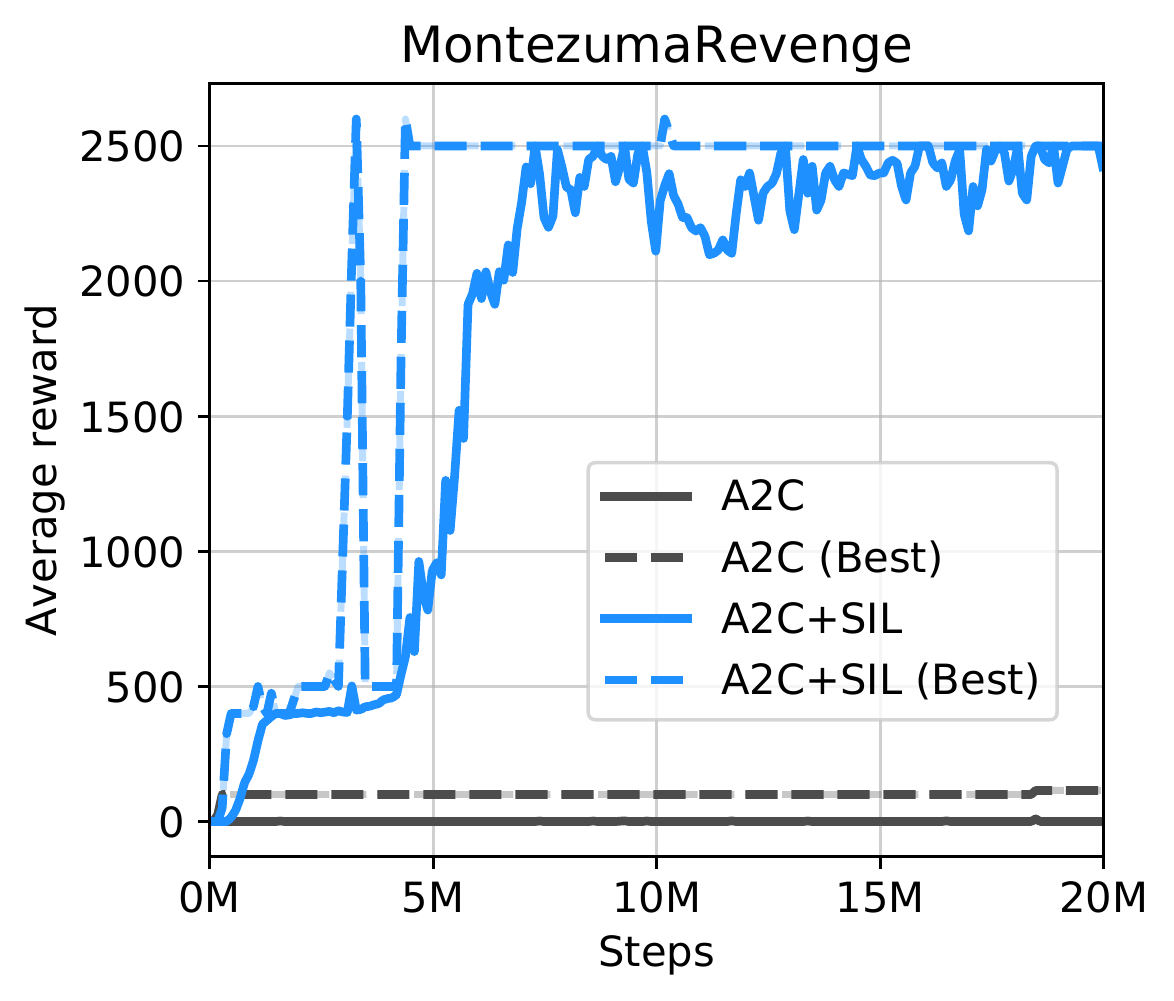} 
   	\end{subfigure}
   	\vskip -0.07in
	\caption{Learning curves on Montezuma's Revenge. (Left) The agent needs to pick up the key in order to open the door. Picking up the key gives a small reward. (Right) The baseline (A2C) often picks up the key as shown by the best episode reward in 100K steps (A2C (Best)), but it fails to consistently reproduce such an experience. In contrast, self-imitation learning (A2C+SIL) quickly learns to pick up the key as soon as the agent experiences it, which leads to the next source of reward (door).} 
	\vskip -0.15in
	\label{fig:intro}
\end{figure}

This paper studies how exploiting the agent's past experiences improves learning in RL. More specifically, we hypothesize that learning to reproduce past good experiences can indirectly lead to deeper exploration depending on the domain. 
A simple example of how this can occur can be seen 
through our results on an example Atari game, Montezuma's Revenge (see Figure~\ref{fig:intro}). In this domain, the first and more proximal source of reward is obtained by picking up the key. Obtaining the key is a precondition of the second and more distal source of reward (i.e., opening the door with the key). Many existing methods occasionally generate experiences that pick up the key and obtain the first reward, but fail to exploit these experiences often enough to learn how to open the door by exploring after picking up the key. Thus, they end up with a poor policy (see A2C in Figure~\ref{fig:intro}).
On the other hand, by exploiting the experiences that pick up the key, the agent is able to explore onwards from the state where it has the key to successfully learn how to open the door (see A2C+SIL in Figure~\ref{fig:intro}). 
Of course, this sort of exploitation can also hurt performance in problems where there are proximal distractor rewards and repeated exploitation of such rewards does not help in learning about more distal and higher rewards; in other words, these two aspects may both be present. 
In this paper we will empirically investigate many different domains to see how exploiting past experiences can be beneficial for learning agents. 

The main contributions of this paper are as follows: (1) To study how exploiting past good experiences affects learning, we propose a \textit{Self-Imitation Learning} (SIL) algorithm which learns to imitate the agent's own past good decisions. In brief, the SIL algorithm stores experiences in a replay buffer, learns to imitate state-action pairs in the replay buffer only when the return in the past episode is greater than the agent's value estimate. (2) We provide a theoretical justification of the SIL objective by showing that the SIL objective is derived from the lower bound of the optimal Q-function. (3) The SIL algorithm is very simple to implement and can be applied to any actor-critic architecture. (4) We demonstrate that SIL combined with advantage actor-critic (A2C) is competitive to the state-of-the-art count-based exploration actor-critic methods (e.g., Reactor-PixelCNN~\citep{ostrovski2017count}) on several hard exploration Atari games~\citep{bellemare13arcade}; SIL also improves the overall performance of A2C across 49 Atari games. Finally, SIL improves the performance of proximal policy optimization (PPO) on MuJoCo continuous control tasks~\citep{brockman2016openai,Todorov2012MuJoCoAP}, demonstrating that SIL may be generally applicable to any actor-critic architecture.

\cutsectionup
\section{Related Work} \label{sec:related-work}
\cutsectiondown

\paragraph{Exploration}
There has been a long history of work on improving exploration in RL, including recent work that can scale up to large state spaces~\cite{stadie2015incentivizing,osband2016deep,bellemare2016unifying,ostrovski2017count}. Many existing methods use some notion of curiosity or uncertainty as a signal for exploration~\citep{Schmidhuber1991AdaptiveCA,strehl2008analysis}. 
In contrast, this paper focuses on exploiting past good experiences for better exploration. Though the role of exploitation for exploration has been discussed~\citep{thrun1992therole}, prior work has mostly considered exploiting what the agent has learned, whereas we consider exploiting what the agent has experienced, but has not yet learned.

\cutparagraphup
\paragraph{Episodic control}
Episodic control~\citep{lengyel2008hippocampal} can be viewed as an extreme way of exploiting past experiences in the sense that the agent repeats the same actions that gave the best outcome in the past. MFEC~\citep{blundell2016model} and NEC~\citep{pritzel2017neural} scaled up this idea to complex domains. However, these methods are slow during test-time because the agent needs to retrieve relevant states for each step and may generalize poorly as the resulting policy is non-parametric.

\cutparagraphup
\paragraph{Experience replay}
Experience replay~\citep{Lin1992SelfImprovingRA} is a natural way of exploiting past experiences for parametric policies. Prioritized experience replay~\citep{Moore1992MemoryBasedRL,Schaul2015PrioritizedER} proposed an efficient way of learning from past experiences by prioritizing them based on temporal-difference error. Our self-imitation learning also prioritizes experiences based on the full episode rewards. Optimality tightening~\citep{He2016LearningTP} introduced an objective based on the lower/upper bound of the optimal Q-function, which is similar to a part of our theoretical result. These recent advances in experience replay have focused on value-based methods such as Q-learning, and are not easily applicable to actor-critic architectures. %On the other hand, this paper proposes a new way of replaying experience for actor-critic architectures.

\cutparagraphup
\paragraph{Experience replay for actor-critic}
In fact, actor-critic framework~\citep{sutton1999policy,konda2000actor} can also utilize experience replay. Many existing methods are based on off-policy policy evaluation~\citep{Precup2001OffPolicyTD,Precup2000EligibilityTF}, which involves importance sampling. For example, ACER~\citep{Wang2016SampleEA} and Reactor~\citep{Gruslys2017TheRA} use Retrace~\citep{Munos2016SafeAE} to evaluate the learner from the behavior policy. Due to importance sampling, this approach may not benefit much from the past experience if the policy in the past is very different from the current policy. Although DPG~\citep{silver2014deterministic,lillicrap2015continuous} performs experience replay without importance sampling, it is limited to continuous control. Our self-imitation learning objective does not involve importance sampling and is applicable to both discrete and continuous control.

\cutparagraphup
\paragraph{Connection between policy gradient and Q-learning}
The recent studies on the relationship between policy gradient and Q-learning have shown that entropy-regularized policy gradient and Q-learning are closely related or even equivalent depending on assumptions~\citep{Nachum2017BridgingTG,ODonoghue2016CombiningPG,Schulman2017EquivalenceBP,Haarnoja2017ReinforcementLW}. %For example, \citet{Schulman2017EquivalenceBP} showed that entropy-regularized policy gradient can be viewed as soft Q-learning in the entropy-regularized RL framework. Our theoretical justification is based on this discovery. 
Our application of self-imitation learning to actor-critic (A2C+SIL) can be viewed as an instance of PGQL~\citep{ODonoghue2016CombiningPG} in that we perform Q-learning on top of actor-critic architecture (see Section~\ref{sec:theory}). Unlike Q-learning in PGQL, however, we use the proposed lower bound Q-learning to exploit good experiences.

\cutparagraphup
\paragraph{Learning from imperfect demonstrations}
A few studies have attempted to learn from imperfect demonstrations, such as DQfD~\citep{hester2018deep}, Q-filter~\citep{Nair2017OvercomingEI}, and normalized actor-critic~\citep{xu2017learning}. Our self-imitation learning has a similar flavor in that the agent learns from imperfect demonstrations. %The objective function of self-imitation learning is similar to that of normalized actor-critic, but we use the Monte-Carlo return instead of 1-step transition. 
However, we treat the agent's own experiences as demonstrations without using expert demonstrations. Although a similar idea has been discussed for program synthesis~\citep{liang2016neural,abolafia2018neural}, this prior work used classification loss without justification. On the other hand, we propose a new objective, provide a theoretical justification, and systematically investigate how it drives exploration in RL.

\cutsectionup
\section{Self-Imitation Learning} \label{sec:method} 
\cutsectiondown
\begin{algorithm}[tb]
\caption{Actor-Critic with Self-Imitation Learning}
\label{alg:sil}
\begin{algorithmic}
\STATE Initialize parameter $\theta$
\STATE Initialize replay buffer $\mathcal{D} \gets \emptyset$
\STATE Initialize episode buffer $\mathcal{E} \gets \emptyset$
\FOR{each iteration}
	\STATE \textbf{\textit{\# Collect on-policy samples}}
	\FOR{each step}
		\STATE Execute an action $s_t,a_t,r_t,s_{t+1} \sim \pi_\theta(a_t|s_t)$
		\STATE Store transition $\mathcal{E} \gets \mathcal{E} \cup \left\{(s_t, a_t, r_t)\right\}$
	\ENDFOR
	\IF{$s_{t+1}$ is terminal}
		\STATE \textbf{\textit{\# Update replay buffer}}
		\STATE Compute returns $R_t=\sum^{\infty}_{k} \gamma^{k-t}r_k$ for all $t$ in $\mathcal{E}$			
		\STATE $\mathcal{D} \gets \mathcal{D} \cup \left\{(s_t, a_t, R_t)\right\}$ for all $t$ in $\mathcal{E}$
		\STATE Clear episode buffer $\mathcal{E} \gets \emptyset$
		%\STATE Keep only $K$-best episodes in $\mathcal{D}$ 	
	\ENDIF
	\STATE \textbf{\textit{\# Perform actor-critic using on-policy samples}}
	\STATE $\theta \gets \theta - \eta \nabla_\theta \mathcal{L}^{a2c}$ \hfill (Eq.~\ref{eq:a2c-loss})
    \STATE \textbf{\textit{\# Perform self-imitation learning}}
    \FOR{$m=1$ to $M$}
	    \STATE Sample a mini-batch $\{(s,a,R)\}$ from $\mathcal{D}$
		\STATE $\theta \gets \theta - \eta \nabla_\theta \mathcal{L}^{sil}$ \hfill (Eq.~\ref{eq:sil-loss})
	\ENDFOR
\ENDFOR
\vskip -0.1in
\end{algorithmic}
\end{algorithm}
%\yijie{line 116~121 in Algorithm 1, what's the relation of $n$ and $t$??}
%\yijie{line 127, how about "keep only state-action-return pair from $K$-best eposodes in $\mathcal{D}$"}

The goal of self-imitation learning (SIL) is to imitate the agent's past good experiences in the actor-critic framework. To this end, we propose to store past episodes with cumulative rewards in a replay buffer: $\mathcal{D}=\{(s_t,a_t,R_t)\}$, where $s_t,a_t$ are a state and an action at time-step $t$, and $R_t=\sum^{\infty}_{k=t}\gamma^{k-t} r_k$ is the discounted sum of rewards with a discount factor $\gamma$. To exploit only good state-action pairs in the replay buffer, we propose the following off-policy actor-critic loss:
\begin{align}
\mathcal{L}^{sil} &= \mathbb{E}_{s,a,R \in \mathcal{D}}\left[\mathcal{L}^{sil}_{policy} + \beta^{sil} \mathcal{L}^{sil}_{value} \right]
\label{eq:sil-loss}
\\
\mathcal{L}^{sil}_{policy} & = - \log \pi_\theta(a|s) \left(R - V_\theta(s)\right)_+ 
\label{eq:sil-policy}
\\ 
\mathcal{L}^{sil}_{value} & =  \frac12 \left\Vert (R - V_\theta(s))_+ \right\Vert^2
\label{eq:sil-value}
\end{align}
where $(\cdot)_+ = \max(\cdot, 0)$, and $\pi_\theta, V_\theta(s)$ are the policy (i.e., actor) and the value function parameterized by $\theta$. $\beta^{sil} \in \mathbb{R}^+$ is a hyperparameter for the value loss.

Note that $\mathcal{L}^{sil}_{policy}$ can be viewed as policy gradient using the value $V_\theta(s)$ as the state-dependent baseline except that we use the off-policy Monte-Carlo return ($R$) instead of on-policy return. $\mathcal{L}^{sil}_{policy}$ can also be interpreted as cross entropy loss (i.e., classification loss for discrete action) with sample weights proportional to the gap between the return and the agent's value estimate ($R-V_\theta$). If the return in the past is greater than the agent's value estimate ($R>V_\theta$), the agent learns to choose the action chosen in the past in the given state. Otherwise ($R\leq V_\theta$), and such a state-action pair is not used to update the parameter due to the $(\cdot)_+$ operator. This encourages the agent to imitate its own decisions in the past only when such decisions resulted in larger returns than expected. $\mathcal{L}^{sil}_{value}$ updates the value estimate towards the off-policy return $R$. 

\paragraph{Prioritized Replay} The proposed self-imitation learning objective $\mathcal{L}^{sil}$ is based on our theoretical result discussed in Section~\ref{sec:theory}. In theory, the replay buffer ($\mathcal{D}$) can be any trajectories from any policies. However, only \textit{good} state-action pairs that satisfy $R>V_\theta$ can contribute to the gradient during self-imitation learning (Eq.~\ref{eq:sil-loss}). Therefore, in order to get many state-action pairs that satisfy $R>V_\theta$, we propose to use the prioritized experience replay~\citep{Schaul2015PrioritizedER}. More specifically, we sample transitions from the replay buffer using the clipped advantage $(R-V_\theta(s))_+$ as priority (i.e., sampling probability is proportional to $(R-V_\theta(s))_+$). This naturally increases the proportion of \textit{valid} samples that satisfy the constraint $(R-V_\theta(s))_+$ in SIL objective and thus contribute to the gradient. 

\cutparagraphup
\paragraph{Advantage Actor-Critic with SIL (A2C+SIL)} 
Our self-imitation learning can be combined with any actor-critic method. In this paper, we focus on the combination of advantage actor-critic (A2C)~\citep{mnih2016asynchronous}  and self-imitation learning (A2C+SIL), as described in Algorithm~\ref{alg:sil}. The objective of A2C ($\mathcal{L}^{a2c}$) is given by~\citep{mnih2016asynchronous}:
\begin{align}
\mathcal{L}^{a2c} &= \mathbb{E}_{s,a \sim \pi_\theta}\left[\mathcal{L}^{a2c}_{policy} + \beta^{a2c} \mathcal{L}^{a2c}_{value} \right] 
\label{eq:a2c-loss}
\\
\mathcal{L}^{a2c}_{policy} & = - \log\pi_\theta(a_t|s_t)(V^n_t - V_\theta(s_t))  - \alpha \mathcal{H}^{\pi_\theta}_t
\label{eq:a2c-policy}
\\
\mathcal{L}^{a2c}_{value} & =  \frac12 \left\Vert V_\theta(s_t) - V^n_t \right\Vert^2 
\label{eq:a2c-value}
\end{align}
where $\mathcal{H}^{\pi}_t=-\sum_{a}\pi(a|s_t)\log \pi(a|s_t)$ denotes the entropy in simplified notation, and $\alpha$ is a weight for entropy regularization. $V^n_t = \sum^{n-1}_{d=0} \gamma^{d}r_{t+d} + \gamma^n V_\theta(s_{t+n})$ is the $n$-step bootstrapped value. 

To sum up, A2C+SIL performs both on-policy A2C update ($\mathcal{L}^{a2c}$) and self-imitation learning from the replay buffer $M$ times ($\mathcal{L}^{sil}$) to exploit past good experiences. A2C+SIL is relatively simple to implement as it does not involve importance sampling.

\cutsectionup
\section{Theoretical Justification} \label{sec:theory}
\cutsectiondown
In this section, we justify the following claim.
\begin{claim}
The self-imitation learning objective ($\mathcal{L}^{sil}$ in Eq.~\ref{eq:sil-loss}) can be viewed as an implementation of lower-bound-soft-Q-learning (Section~\ref{sec:lower-bound-soft-q-learning}) under the entropy-regularized RL framework. 
\end{claim}
To show the above claim, we first introduce the entropy-regularized RL~\citep{Haarnoja2017ReinforcementLW} in Section~\ref{sec:background}. Section~\ref{sec:lower-bound-soft-q-learning} introduces \textit{lower-bound-soft-Q-learning}, an off-policy Q-learning algorithm, which learns the optimal action-value function from good state-action pairs. Section~\ref{sec:connection} proves the above claim by showing the equivalence between self-imitation learning and lower-bound-soft-Q-learning. Section~\ref{sec:a2c-sil} further discusses the relationship between A2C and self-imitation learning.

\cutsubsectionup
\subsection{Entropy-Regularized Reinforcement Learning} \label{sec:background}
\cutsubsectiondown
The goal of entropy-regularized RL is to learn a stochastic policy which maximizes the entropy of the policy as well as the $\gamma$-discounted sum of rewards~\citep{Haarnoja2017ReinforcementLW,Ziebart2008MaximumEI}:
\begin{align}
\pi^{*} = \text{argmax}_\pi \mathbb{E}_{\pi}\left[\sum_{t=0}^{\infty} \gamma^{t} \left(r_t + \alpha \mathcal{H}^{\pi}_t \right) \right]
\label{eq:ent-rl}
\end{align}
where $\mathcal{H}^{\pi}_t=-\log \pi(a_t|s_t)$ is the entropy of the policy $\pi$, and $\alpha \geq 0$ represents the weight of entropy bonus. Intuitively, in the entropy-regularized RL, a policy that has a high entropy is preferred (i.e., diverse actions chosen given the same state).

The optimal \textit{soft} Q-function and the optimal \textit{soft} value function are defined as:
\begin{align}
Q^{*}(s_t,a_t) & = \mathbb{E}_{\pi^{*}}\left[r_t + \sum_{k=t+1}^{\infty} \gamma^{k-t} (r_{k} + \alpha \mathcal{H}^{\pi^{*}}_{k})\right] 
\label{eq:optimal-soft-q}
\\
V^{*}(s_t) & = \alpha \log \sum_{a}\exp \left( Q^{*}(s_t, a) /\alpha \right).
\label{eq:optimal-soft-v}
\end{align}
It is shown that the optimal policy $\pi^{*}$ has the following form (see \citet{ziebart2010modeling,Haarnoja2017ReinforcementLW} for the proof):
\begin{align}
\pi^{*}(a|s) & = \exp((Q^{*}(s,a)-V^{*}(s))/\alpha).
\label{eq:optimal-soft-pi}
\end{align} 
This result provides the relationship among the optimal Q-function, the optimal policy, and the optimal value function, which will be useful in Section~\ref{sec:connection}.

\cutsubsectionup
\subsection{Lower Bound Soft Q-Learning} \label{sec:lower-bound-soft-q-learning}
\cutsubsectiondown
\paragraph{Lower bound of optimal soft Q-value} Let $\pi^{*}$ be an optimal policy in entropy-regularized RL (Eq.~\ref{eq:ent-rl}). It is straightforward that the expected return of any behavior policy $\mu$ can serve as a lower bound of the optimal soft Q-value as follows:
\begin{align}
Q^{*}(s_t, a_t) &= \mathbb{E}_{\pi^{*}}\left[r_t + \sum^{\infty}_{k=t+1} \gamma^{k-t}(r_k + \alpha \mathcal{H}^{\pi^{*}}_k) \right] 
\\
& \geq \mathbb{E}_{\mu}\left[r_t + \sum^{\infty}_{k=t+1} \gamma^{k-t}(r_k + \alpha \mathcal{H}^{\mu}_k) \right], 
\label{eq:q-lower-bound}
\end{align}
because the entropy-regularized return of the optimal policy is always greater or equal to that of any other policies.
\paragraph{Lower bound soft Q-learning}
Suppose that we have full episode trajectories from a behavior policy $\mu$, which consists of state-action-return triples: $\left(s_t,a_t,R_t\right)$ where $R_t=r_t + \sum^\infty_{k=t+1} \gamma^{k-t}(r_{k} + \alpha \mathcal{H}^{\mu}_k)$ is the entropy-regularized return. 
We propose \textit{lower bound soft Q-learning} which updates $Q_\theta(s,a)$ parameterized by $\theta$ towards the optimal soft Q-value as follows ($t$ is omitted for brevity):
\begin{align}
\mathcal{L}^{lb} = \mathbb{E}_{s,a,R \sim \mu} \left[\frac12 \left\Vert (R - Q_\theta(s, a))_+ \right\Vert^2 \right],
\label{eq:lower-bound-q}
\end{align}
where $(\cdot)_+ = \max(\cdot, 0)$. 
Intuitively, we update the Q-value only when the return is greater than the Q-value estimate ($R>Q_\theta(s,a)$). This is justified by the fact that the lower bound (Eq.~\ref{eq:q-lower-bound}) implies that the estimated Q-value is lower than the optimal soft Q-value: $Q^{*}(s,a)\geq R > Q_\theta(s,a)$ when the environment is deterministic. 
Otherwise ($R\leq Q_\theta(s,a)$), such state-action pairs do not provide any useful information about the optimal soft Q-value, so they are not used for training. 
We call this lower-bound-soft-Q-learning as it updates Q-values towards the lower bounds of the optimal Q-values observed from the behavior policy.

\cutsubsectionup
\subsection{Connection between SIL and Lower Bound Soft Q-Learning}  \label{sec:connection}
\cutsubsectiondown
In this section, we derive an equivalent form of lower-bound-soft-Q-learning (Eq.~\ref{eq:lower-bound-q}) for the actor-critic architecture and show a connection to self-imitation learning objective. 

Suppose that we have a parameterized soft Q-function $Q_\theta$. According to the form of optimal soft value function and optimal policy in the entropy-regularized RL (Eq.~\ref{eq:optimal-soft-v}-\ref{eq:optimal-soft-pi}), it is natural to consider the following forms of a value function $V_\theta$ and a policy $\pi_\theta$:
\begin{align}
V_\theta(s) & = \alpha\log\sum_a\exp(Q_\theta(s,a)/\alpha) \\
\pi_\theta(a|s) & = \exp((Q_\theta(s,a)-V_\theta(s))/\alpha).
\end{align}
From these definitions, $Q_\theta$ can be written as:
\begin{align}
Q_\theta(s,a) & = V_\theta(s) + \alpha \log\pi_\theta(a|s).
\label{eq:soft-q-v2}
\end{align}
For convenience, let us define the following: 
\begin{align}
\hat{R}&=R - \alpha\log\pi_\theta(a|s)
\\
\Delta &= R - Q_\theta(s,a) = \hat{R}-V_\theta(s).
\end{align}
By plugging Eq.~\ref{eq:soft-q-v2} into Eq.~\ref{eq:lower-bound-q}, we can derive the gradient estimator of lower-bound-soft-Q-learning for the actor-critic architecture as follows:
\begin{align}
%\nabla_\theta\mathcal{L}^{lb} 
&\nabla_\theta \mathbb{E}_{s,a,R\sim \mu} \left[ \frac{1}{2} \left\Vert (R-Q_\theta(s,a))_+ \right\Vert^2 \right]
\\
&=\mathbb{E} \left[ -\nabla_\theta Q_\theta(s,a)\Delta_+ \right]
\\
&=\mathbb{E}\left[-\nabla_\theta \left(\alpha\log \pi_\theta(a|s) + V_\theta(s)\right)\Delta_+ \right]
\\
&=\mathbb{E}\left[-\alpha \nabla_\theta \log \pi_\theta(a|s)\Delta_+ - \nabla_\theta V_\theta(s)\Delta_+  \right]
\\
&=\mathbb{E}\left[\alpha \nabla_\theta \mathcal{L}^{lb}_{policy} - \nabla_\theta V_\theta(s)\Delta_+  \right]
\\
&=\mathbb{E}\left[\alpha \nabla_\theta \mathcal{L}^{lb}_{policy} - \nabla_\theta V_\theta(s)(R-Q_\theta(s,a))_+  \right]
\\
&=\mathbb{E}\left[\alpha \nabla_\theta \mathcal{L}^{lb}_{policy} - \nabla_\theta V_\theta(s)(\hat{R}-V_\theta(s) )_+  \right]
\\
&=\mathbb{E}\left[\alpha \nabla_\theta \mathcal{L}^{lb}_{policy} + \nabla_\theta \frac12 \left\Vert (\hat{R} - V_\theta(s))_+ \right\Vert^2  \right]
\\
&=\mathbb{E}\left[\alpha \nabla_\theta \mathcal{L}^{lb}_{policy} + \nabla_\theta \mathcal{L}^{lb}_{value}  \right].
\label{eq:off-soft-q}
\end{align}
Each loss term in Eq.~\ref{eq:off-soft-q} is given by:
\begin{align}
\mathcal{L}^{lb}_{policy} & = - \log \pi_\theta(a|s) \left(\hat{R} - V_\theta(s)\right)_+ 
\label{eq:appendix-off-pg}
\\ 
\mathcal{L}^{lb}_{value} & =  \frac12 \left\Vert (\hat{R} - V_\theta(s))_+ \right\Vert^2.
\label{eq:appendix-off-val}
\end{align}
Thus, $\mathcal{L}^{lb}_{policy} = \mathcal{L}^{sil}_{policy}$ and 
$\mathcal{L}^{lb}_{value} = \mathcal{L}^{sil}_{value}$ as $\alpha \rightarrow 0$ (see Eq.~\ref{eq:sil-policy}-\ref{eq:sil-value}). This shows that the proposed self-imitation learning objective $\mathcal{L}^{sil}$ (Eq.~\ref{eq:sil-loss}) can be viewed as a form of lower-bound-soft-Q-learning (Eq.~\ref{eq:lower-bound-q}), but without explicitly optimizing for entropy bonus reward as $\alpha \rightarrow 0$.
Since the lower-bound-soft-Q-learning directly updates the Q-value towards the lower bound of the optimal Q-value, self-imitation learning can be viewed as an algorithm that updates the policy ($\pi_\theta$) and the value ($V_\theta$) directly towards the optimal policy and the optimal value respectively.

\cutsubsectionup
\subsection{Relationship between A2C and SIL} \label{sec:a2c-sil}
Intuitively, A2C updates the policy in the direction of increasing the expected return of the learner policy and enforces consistency between the value and the policy from on-policy trajectories. On the other hand, SIL updates each of them directly towards optimal policies and values respectively from off-policy trajectories. In fact, \citet{Nachum2017BridgingTG,Haarnoja2017ReinforcementLW,Schulman2017EquivalenceBP} have recently shown that entropy-regularized A2C can be viewed as $n$-step online soft Q-learning (or path consistency learning). Therefore, both A2C and SIL objectives are designed to learn the optimal soft Q-function in the entropy-regularized RL framework. 
Thus, we claim that both objectives can be complementary to each other in that they share the same optimal solution as discussed in PGQL~\citep{ODonoghue2016CombiningPG}.

\begin{figure}
	\centering
	\begin{subfigure}{0.98\linewidth}
		\centering
		\raisebox{0.1\height}{\includegraphics[width=0.4\linewidth]{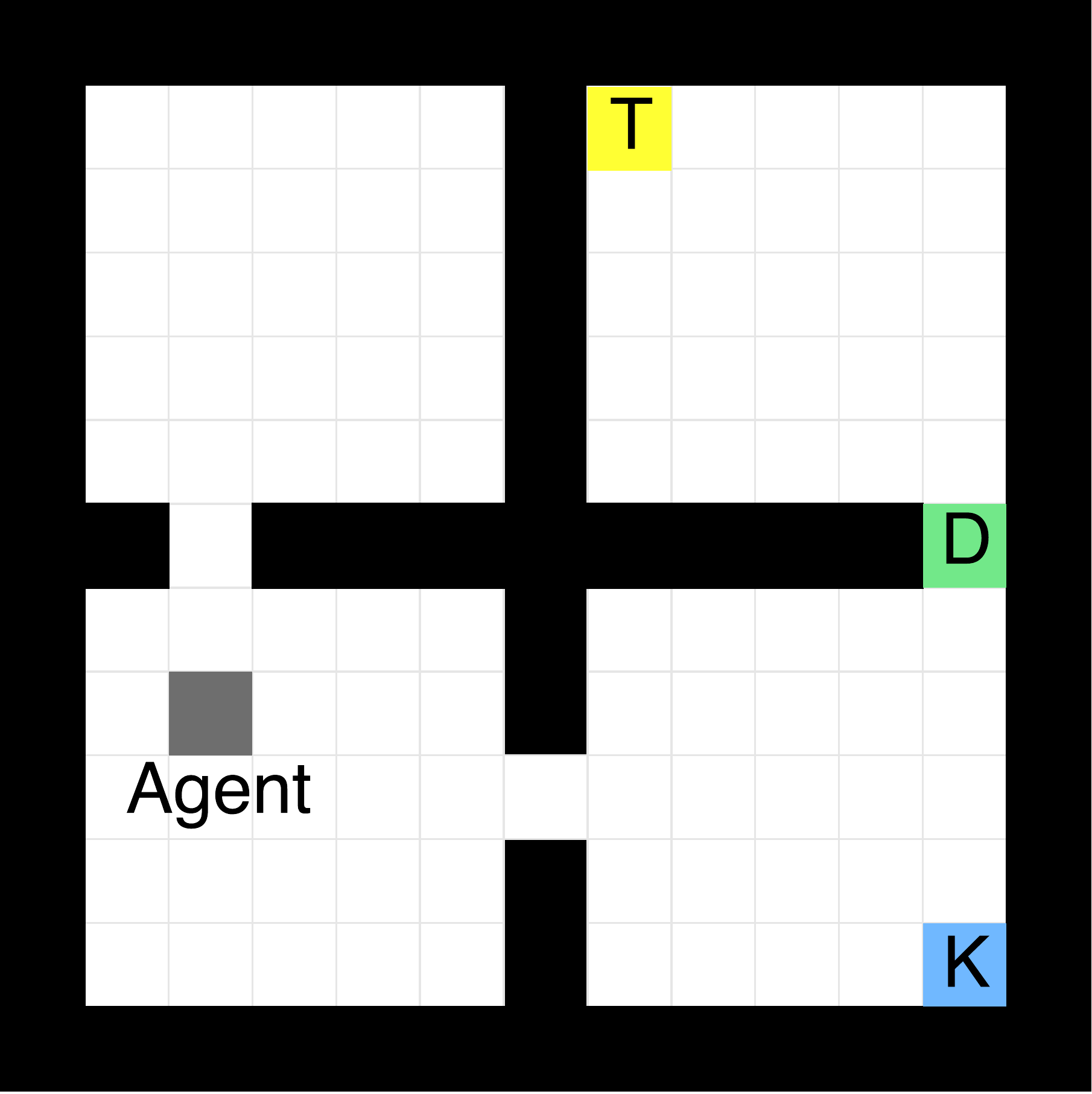}}
		\hskip 0.1in
		\includegraphics[width=0.55\linewidth]{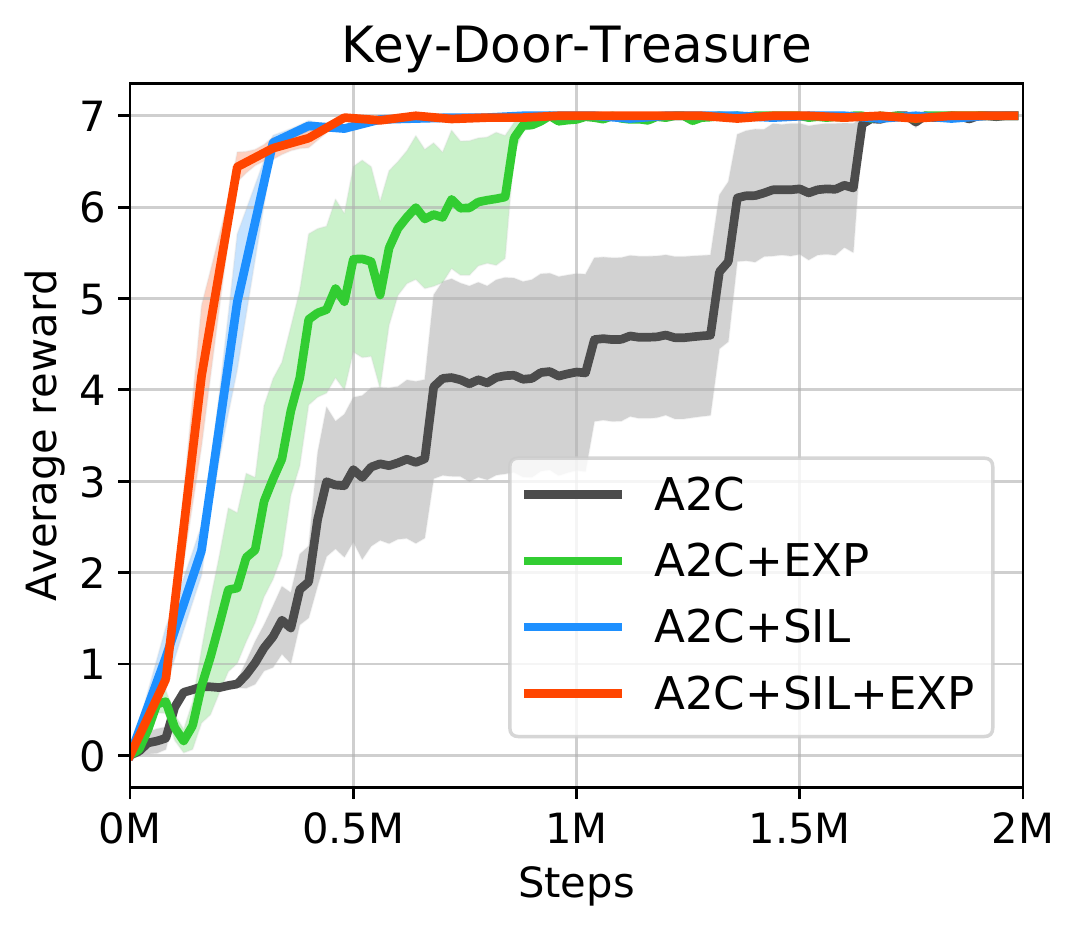}
	\end{subfigure}
	%\\
	\centering
	\begin{subfigure}{0.98\linewidth}
		\centering
		\raisebox{0.1\height}{\includegraphics[width=0.4\linewidth]{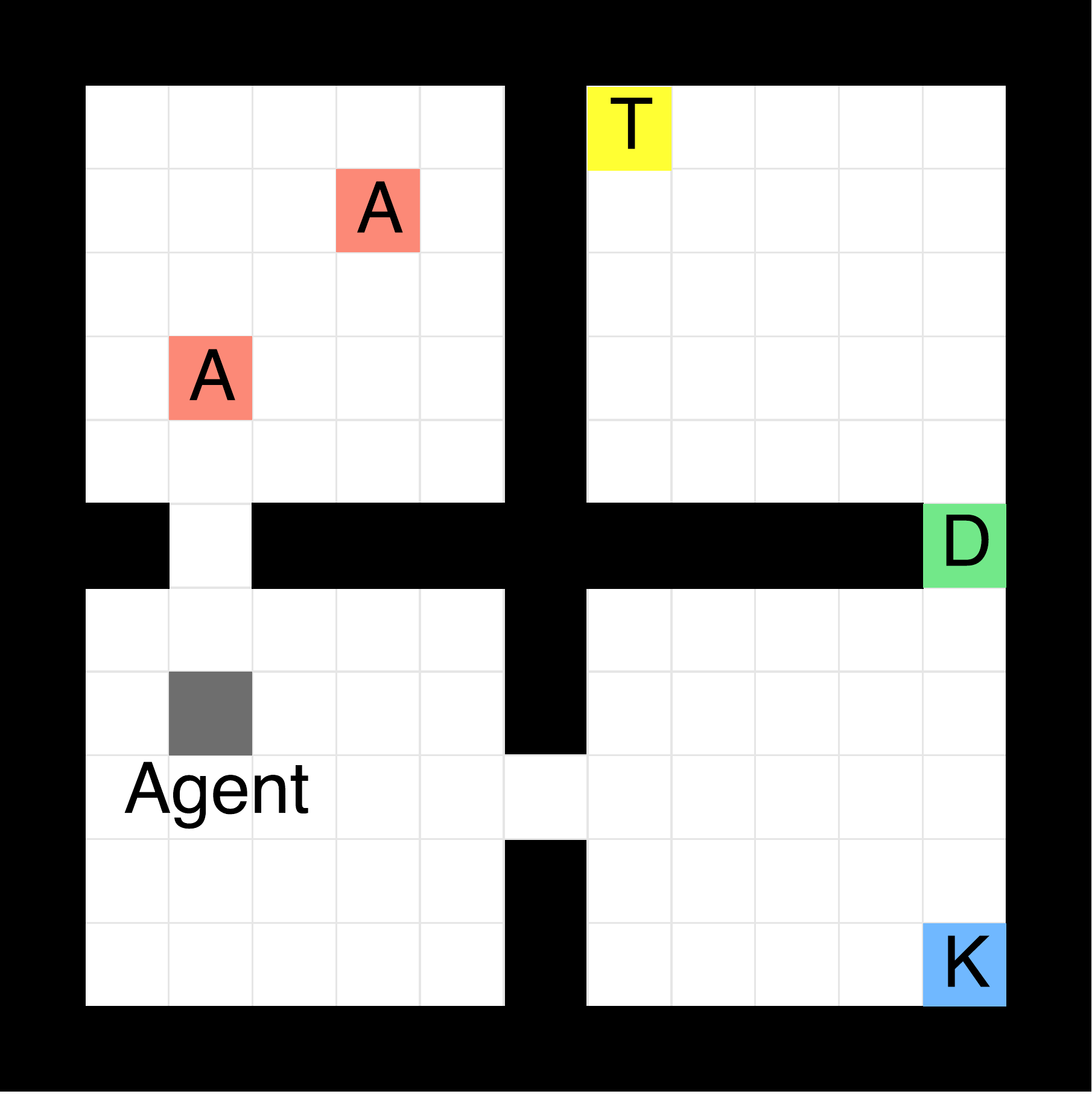}}
		\hskip 0.1in
		\includegraphics[width=0.55\linewidth]{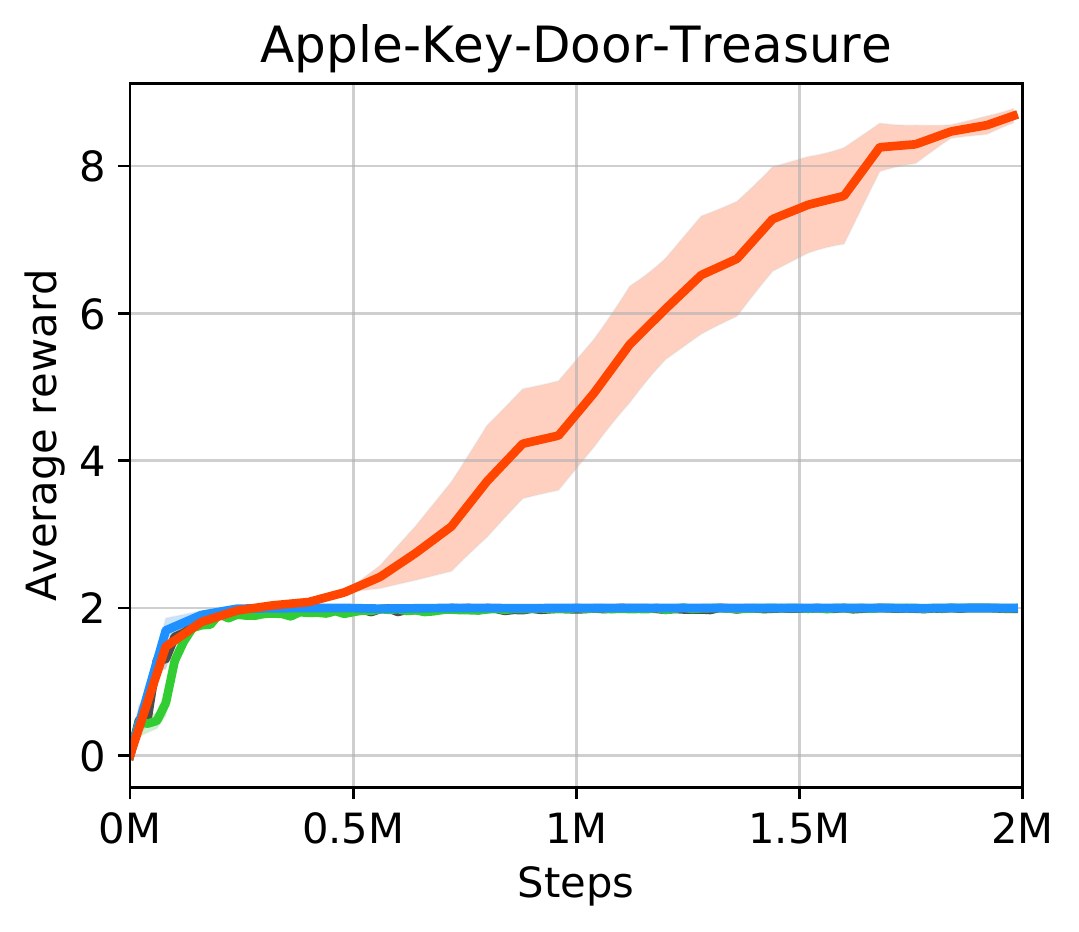}
	\end{subfigure}
	\vskip -0.05in
	\caption{Key-Door-Treasure domain. The agent should pick up the key (K) in order to open the door (D) and collect the treasure (T) to maximize the reward. In the Apple-Key-Door-Treasure domain (bottom), there are two apples (A) that give small rewards (+1). `SIL' and `EXP' represent our self-imitation learning and a count-based exploration method respectively. }
	\label{fig:grid-world}
	\vskip -0.15in
\end{figure}

\cutsectionup
\section{Experiment} \label{sec:experiment}
\cutsectiondown
The experiments are designed to answer the following:
\begin{itemize}[leftmargin=*]
\setlength\itemsep{0em}
\item Is self-imitation learning useful for exploration?
\item Is self-imitation learning complementary to count-based exploration methods?
\item Does self-imitation learning improve the overall performance across a variety of tasks?
\item When does self-imitation learning help and when does it not?
\item Can other off-policy actor-critic methods also exploit good experiences (e.g., ACER~\citep{Wang2016SampleEA})?
\item Is self-imitation learning useful for continuous control and compatible with other learning algorithms like PPO~\citep{Schulman2017ProximalPO}?
\end{itemize}

\begin{figure*}
	\centering
	\includegraphics[width=0.8\linewidth]{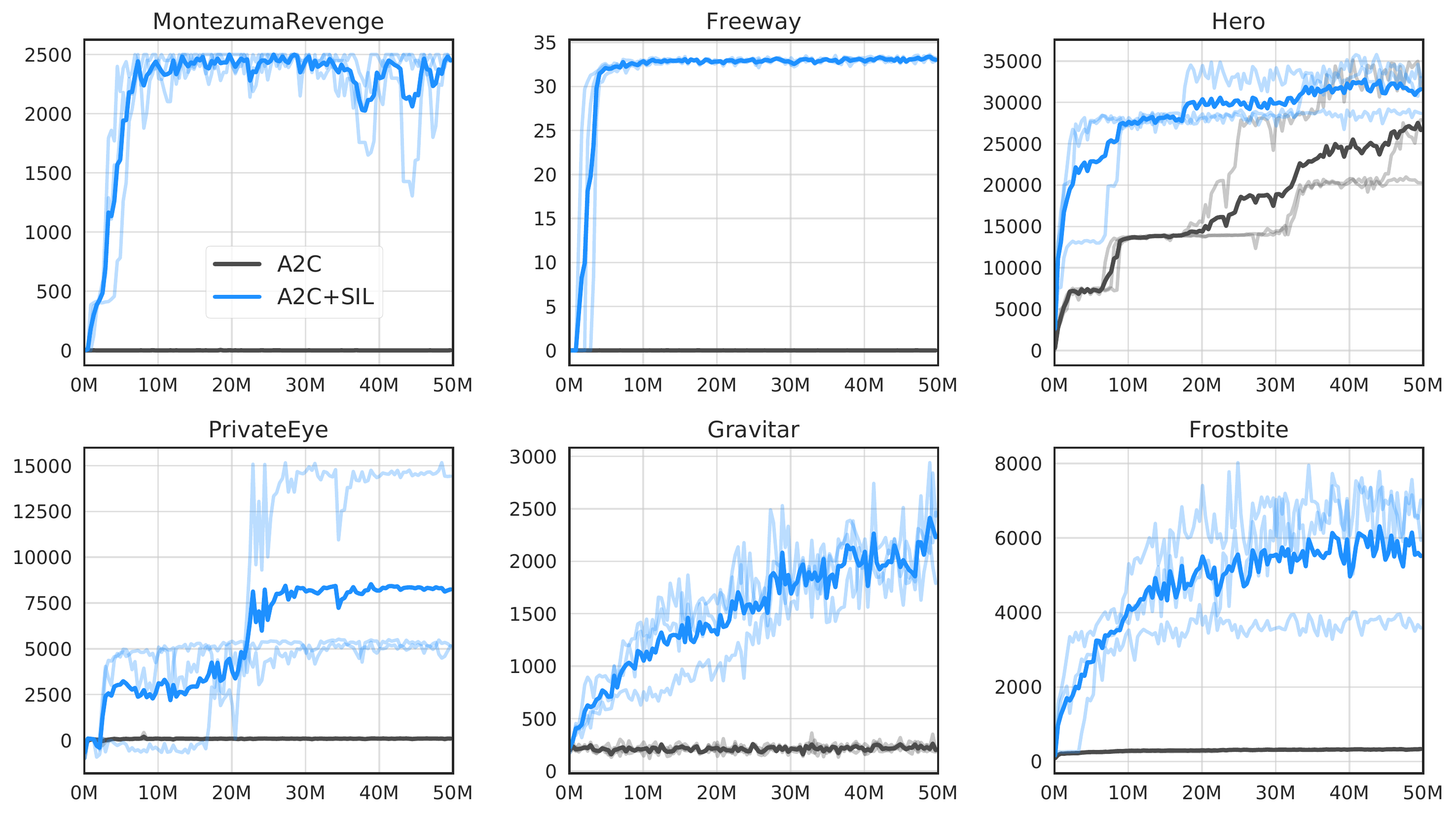}
	\vspace{-0.1in}
	\caption{Learning curves on hard exploration Atari games. X-axis and y-axis represent steps and average reward respectively. }
	\label{fig:hard-exploration}
	\vskip -0.1in
\end{figure*}

\cutsubsectionup
\subsection{Implementation Details}
\cutsubsectiondown
For Atari experiments, we used a 3-layer convolutional neural network used in DQN~\citep{mnih2015human} with last 4 stacked frames as input. We performed 4 self-imitation learning updates per on-policy actor-critic update ($M=4$ in Algorithm~\ref{alg:sil}). Instead of treating losing a life as episode termination as typically done in the previous work, we terminated episodes when the game ends, as it is the true definition of episode termination. 
For MuJoCo experiments, we used an MLP which consists of 2 hidden layers with 64 units as in \citet{Schulman2017ProximalPO}. We performed 10 self-imitation learning updates per each iteration (batch).
More details of the network architectures and hyperparameters are described in \supp{}. Our implementation is based on OpenAI's baseline implementation~\citep{baselines}.\footnote{The code is available on \url{https://github.com/junhyukoh/self-imitation-learning}. }

\cutsubsectionup
\subsection{Key-Door-Treasure Domain}
\cutsubsectiondown
To investigate how self-imitation learning is useful for exploration and whether it is complementary to count-based exploration method, we compared different methods on a grid-world domain, as illustrated in Figure~\ref{fig:grid-world}. More specifically, we implemented a count-based exploration method~\citep{strehl2008analysis} that gives an exploration bonus reward: $r_{exp} = \beta / \sqrt{N(s)}$, where $N(s)$ is the visit count of state $s$ and $\beta$ is a hyperparameter. We also implemented a combination with self-imitation learning shown as `A2C+SIL+EXP' in Figure~\ref{fig:grid-world}.
%A2C trained with this exploration bonus reward is denoted by `A2C+EXP' in Figure~\ref{fig:grid-world}. 

In the first domain (Key-Door-Treasure), the chance of picking up the key followed by opening the door and obtaining the treasure is low due to the sequential dependency between them. We found that the baseline A2C tends to get stuck at a sub-optimal policy that only opens the door for a long time. A2C+EXP learns faster than A2C because exploration bonus encourages the agent to collect the treasure more often. Interestingly, A2C+SIL and A2C+SIL+EXP learn most quickly. We observed that once the agent opens the door with the key by chance, our SIL helps exploit such good experiences and quickly learns to open the door with the key. This increases the chance of getting the next reward (i.e., treasure) and helps learn the optimal policy. This is an example showing that self-imitation learning can drive deep exploration. 

In the second domain (Apple-Key-Door-Treasure), collecting apples near the agent's initial location makes it even more challenging for the agent to learn the optimal policy, which collects all of the objects within the time limit (50 steps). In this domain, many agents learned a sub-optimal policy that only collects two apples as shown in Figure~\ref{fig:grid-world}. On the other hand, only A2C+SIL+EXP consistently learned the optimal policy because count-based exploration increases the chance of collecting the treasure, while self-imitation learning can quickly exploit such a good experience as soon as the agent collects it. This result shows that self-imitation learning and count-based exploration methods can be complementary to each other. This result also suggests that while exploration is important for increasing the chance/frequency of getting a reward, it is also important to exploit such rare experiences to learn a policy to consistently achieve it especially when the reward is sparse.

\begin{table}
\caption{Comparison to count-based exploration actor-critic agents on hard exploration Atari games. A3C+ and Reactor+ correspond to A3C-CTS~\citep{bellemare2016unifying} and Reactor-PixelCNN respectively~\citep{ostrovski2017count}. SimHash represents TRPO-AE-SimHash~\citep{tang2017exploration}. $^{\dagger}$Numbers are taken from plots. }
\label{tab:exp}
\begin{center}
\begin{small}
\begin{sc}
\setlength{\tabcolsep}{3pt}
\begin{tabular}{lcccc}
\toprule
 & \textbf{A2C+SIL} & A3C+ & Reactor+$^{\dagger}$ & SimHash  \\
\midrule
Montezuma & \textbf{2500} & 273 & 100 & 75 \\
Freeway  & \textbf{34} & 30 & 32 & 33 \\
Hero  & \textbf{33069} & 15210 & 28000 & N/A \\
PrivateEye  & \textbf{8684} & 99 & 200 & N/A \\
Gravitar & \textbf{2722} & 239 & 1600 & 482 \\
Frostbite  & \textbf{6439} & 352 & 4800 & 5214 \\
Venture  & 0 & 0 & \textbf{1400} & 445 \\
\bottomrule
\end{tabular}
\end{sc}
\end{small}
\end{center}
\vskip -0.2in
\end{table}
\cutsubsectionup
\subsection{Hard Exploration Atari Games} \label{sec:exp-hard}
\cutsubsectiondown
We investigated how useful our self-imitation learning is for several hard exploration Atari games on which recent advanced exploration methods mainly focused. Figure~\ref{fig:hard-exploration} shows that A2C with our self-imitation learning (A2C+SIL) outperforms A2C on six hard exploration games. A2C failed to learn a better-than-random policy, except for Hero, whereas our method learned better policies and achieved human-level performances on Hero and Freeway. We observed that even a random exploration occasionally leads to a positive reward on these games, and self-imitation learning helps exploit such an experience to learn a good policy from it. This can drive deep exploration when the improved policy gets closer to the next source of reward. This result supports our claim that exploiting past experiences can often help exploration.

We further compared our method against the state-of-the-art count-based exploration actor-critic agents (A3C-CTS~\citep{bellemare2016unifying}, Reactor-PixelCNN~\citep{ostrovski2017count}, and SimHash~\citep{tang2017exploration}). These methods learn a density model of the observation or a hash function and use it to compute pseudo visit count, which is used to compute an exploration bonus reward. Even though our method does not have an explicit exploration bonus that encourages exploration, we were curious how well our self-imitation learning approach performs compared to these exploration approaches.

Interestingly, Table~\ref{tab:exp} shows that A2C with our self-imitation learning (A2C+SIL) achieves better results on 6 out of 7 hard exploration games without any technique that explicitly encourages exploration. This result suggests that it is important to exploit past good experiences as well as efficiently explore the environment to drive deep exploration.

On the other hand, we found that A2C+SIL never receives a positive reward on Venture during training. This makes it impossible for our method to learn a good policy because there is no good experience to exploit, whereas one of the count-based exploration methods (Reactor-PixelCNN) achieves a better performance, because the agent is encouraged to explore different states even in the absence of reward signal from the environment. This result suggests that an advanced exploration method is essential in such environments where a random exploration never generates a good experience within a reasonable amount of time. 
Combining self-imitation learning with state-of-the-art exploration methods would be an interesting future work.

\begin{figure}
	\centering
	\includegraphics[width=0.98\linewidth]{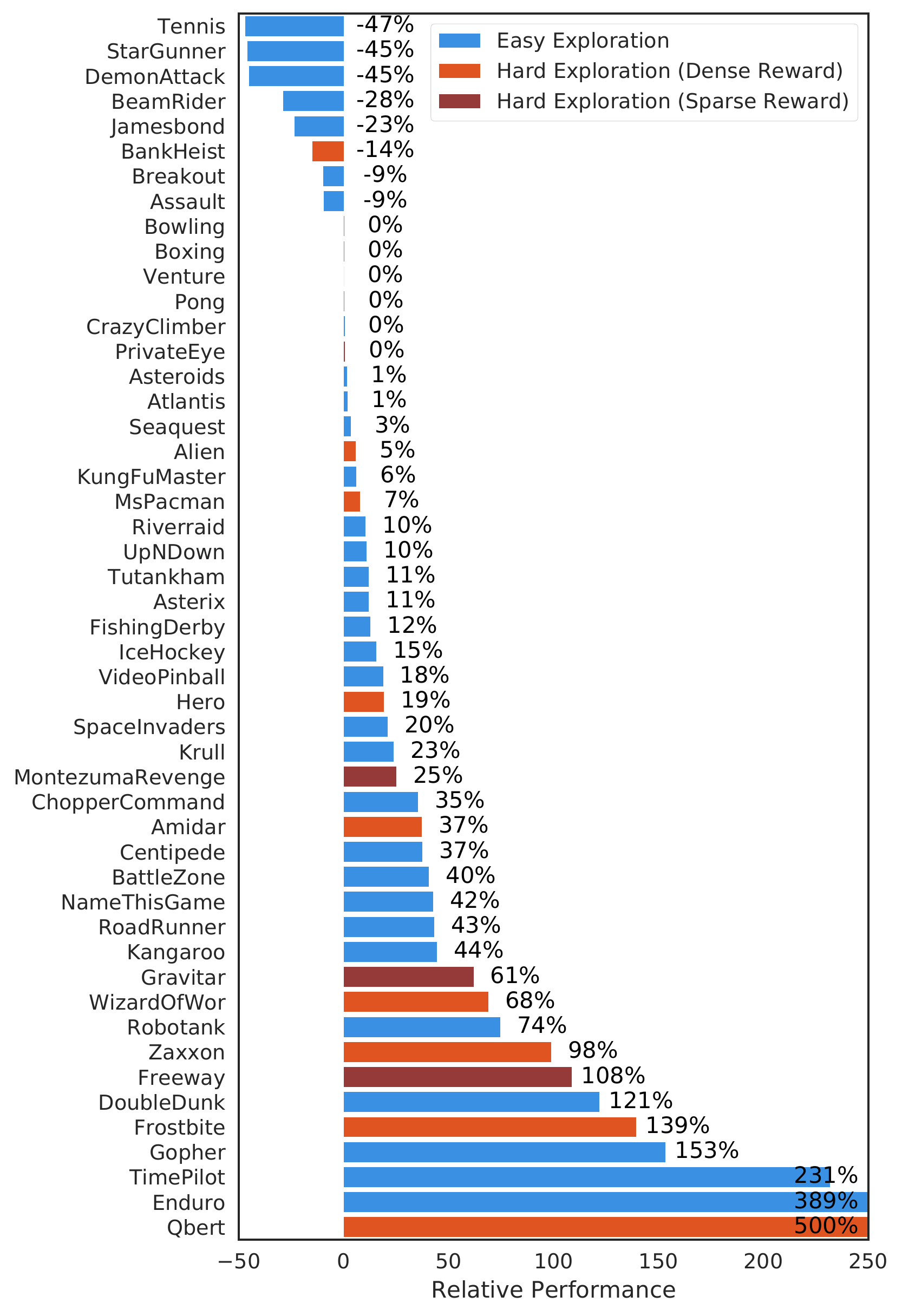}
	\vskip -0.1in
	\caption{Relative performance of A2C+SIL over A2C.}
	\label{fig:relative}
	\vskip -0.1in
\end{figure}

%\begin{figure*}
%	\centering
%	\includegraphics[width=0.98\linewidth]{figures/atari_prioritized_horizontal_relative_score_A2C+SIL.pdf}
%	\vskip -0.1in
%	\caption{Relative performance of A2C+SIL over A2C.}
%	\label{fig:relative}
%	\vskip -0.1in
%\end{figure*}

\cutsubsectionup
\subsection{Overall Performance on Atari Games} \label{sec:exp-overall}
\cutsubsectiondown
To see how useful self-imitation learning is across various types of environments, we evaluated our self-imitation learning method on 49 Atari games. It turns out that our method (A2C+SIL) significantly outperforms A2C in terms of median human-normalized score as shown in Table~\ref{tab:atari}. Figure~\ref{fig:relative} shows the relative performance of A2C+SIL compared to A2C using the measure proposed by~\citet{Wang2016DuelingNA}.
It is shown that our method (A2C+SIL) improves A2C on 35 out of 49 games in total and 11 out of 14 hard exploration games defined by~\citet{bellemare2016unifying}. It is also shown that A2C+SIL performs significantly better on many easy exploration games such as Time Pilot as well as hard exploration games. We observed that there is a certain learning stage where the agent suddenly achieves a high score by chance on such games, and our self-imitation learning exploits such experiences as soon as the agent experiences them.

On the other hand, we observed that our method often learns faster at the early stage of learning, but sometimes gets stuck at a sub-optimal policy on a few games, such as James Bond and Star Gunner. This suggests that excessive exploitation at the early stage of learning can hurt the performance. We found that reducing the number of SIL updates per iteration or using a small weight for the SIL objective in a later learning stage indeed resolves this issue and even improve the performance on such games, though the reported numbers are based on the single best hyperparameter. Thus, automatically controlling the degree of self-imitation learning would be an interesting future work.

\begin{table}
\caption{Performance of agents on 49 Atari games after 50M steps (200M frames) of training. `ACPER' represents A2C with prioritized replay using ACER objective. `Median' shows median of human-normalized scores. `$>$Human' shows the number of games where the agent outperforms human experts. }
\label{tab:atari}
\begin{center}
\begin{small}
\begin{sc}
\begin{tabular}{lccc}
\toprule
Agent & Median & $>$Human \\
\midrule
A2C & 96.1\% & 23 \\
ACPER & 46.8\% & 18 \\
\midrule
\textbf{A2C+SIL} & \textbf{138.7\%} & \textbf{29} \\
%A2C & 96.1\% & 36 & 31 & 23 \\
%ACPER & 46.8\% & 36 & 23 & 18 \\
%\textbf{A2C+SIL} & \textbf{138.7\%} & \textbf{44} & \textbf{39} & \textbf{29} \\
%\midrule
\bottomrule
\end{tabular}
\end{sc}
\end{small}
\end{center}
\vskip -0.3in
\end{table}

\begin{figure*}
	\centering
	\includegraphics[width=0.99\linewidth]{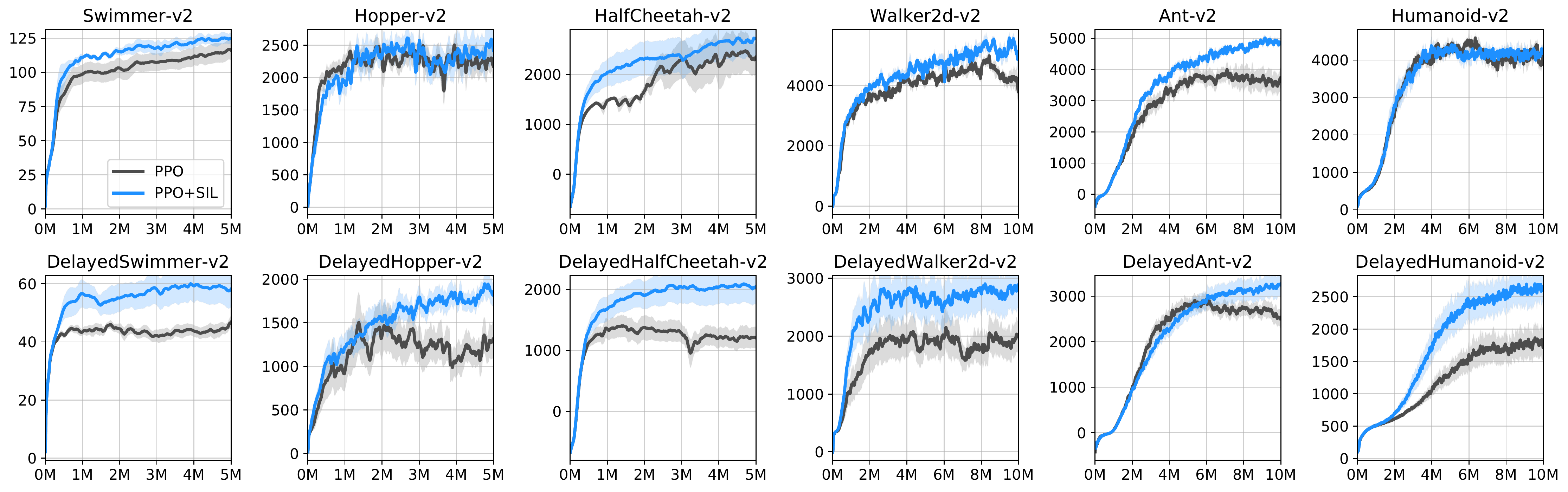}
	\vspace{-0.1in}
	\caption{Performance on OpenAI Gym MuJoCo tasks (top row) and delayed-reward versions of them (bottom row). The learning curves are averaged over 10 random seeds.}
	\label{fig:mujoco}
	\vskip -0.1in
\end{figure*}

\cutsubsectionup
\subsection{Effect of Lower Bound Soft Q-Learning}
\cutsubsectiondown
A natural question is whether existing off-policy actor-critic methods can also benefit from past good experiences by exploiting them. To answer this question, we trained ACPER (A2C with prioritized experience replay) which performs off-policy actor-critic update proposed by ACER~\citep{Wang2016SampleEA} by using the same prioritized experience replay as ours, which uses $(R-V_\theta)_+$ as sampling priority. ACPER can also be viewed as the original ACER with our proposed prioritized experience replay. 

Table~\ref{tab:atari} shows that ACPER performs much worse than our A2C+SIL and is even worse than A2C. We observed that ACPER also benefits from good episodes on a few hard exploration games (e.g., Freeway) but was very unstable on many other games. 

We conjecture that this is due to the fact that the ACER objective has an importance weight term ($\pi(a|s) / \mu(a|s)$). This approach may not benefit much from the good experiences in the past if the current policy deviates too much from the decisions made in the past. On the other hand, the proposed self-imitation learning objective (Eq.~\ref{eq:sil-loss}) does not have an importance weight and can learn from any behavior, as long as the behavior policy performs better than the learner. This is because our gradient estimator can be interpreted as lower-bound-soft-Q-learning, which updates the parameter directly towards the optimal Q-value regardless of the similarity between the behavior policy and the learner as discussed in Section~\ref{sec:lower-bound-soft-q-learning}.
This result shows that our self-imitation learning objective is suitable for exploiting past good experiences.

\cutsubsectionup
\subsection{Performance on MuJoCo}
\cutsubsectiondown
This section investigates whether self-imitation learning is beneficial for continuous control tasks and whether it can be applied to other types of policy optimization algorithms, such as proximal policy optimization (PPO)~\citep{Schulman2017ProximalPO}. Note that unlike A2C, PPO does not have a strong theoretical connection to our SIL objective. However, we claim that they can still be complementary to each other in that both PPO and SIL try to update the policy and the value towards the optimal policy and value.
To empirically verify this, we implemented PPO+SIL, which updates the parameter using both the PPO algorithm and our SIL algorithm and evaluated it on 6 MuJoCo tasks in OpenAI Gym~\citep{brockman2016openai}. 

The result in Figure~\ref{fig:mujoco} shows that our self-imitation learning improves PPO on Swimmer, Walker2d, and Ant tasks. 
Unlike Atari games, the reward structure in this benchmark is smooth and dense in that the agent always receives a reasonable amount of reward according to its continuous progress. We conjecture that the agent has a relatively low chance to occasionally perform well and learn much faster by exploiting such an experience in this type of domain. 
Nevertheless, the overall improvement suggests that self-imitation learning can be generally applicable to actor-critic architectures and a variety of tasks. 

To verify our conjecture, we further conducted experiments by delaying reward the agent gets from the environment. More specifically, the modified tasks give an accumulated reward after every 20 steps (or when the episode terminates). This makes it more difficult to learn a good policy because the agent does not receive a reward signal for every step. 
The result is shown in the bottom row in Figure~\ref{fig:mujoco}. Not surprisingly, we observed that both PPO and PPO+SIL perform worse on the delayed-reward tasks than themselves on the standard OpenAI Gym tasks. However, it is clearly shown that the gap between PPO+SIL and PPO is larger on delayed-reward tasks compared to standard tasks. Unlike the standard OpenAI Gym tasks where reward is well-designed and dense, we conjecture that the chance of achieving high overall rewards is much low in the delayed-reward tasks. Thus, the agent can benefit more from self-imitation learning because self-imitation learning captures such rare experiences and learn from them. 

\cutsectionup
\section{Conclusion}
\cutsectiondown
In this paper, we proposed self-imitation learning, which learns to reproduce the agent's past good experiences, and showed that self-imitation learning is very helpful on hard exploration tasks as well as a variety of other tasks including continuous control tasks. 
We also showed that a proper level of exploitation of past experiences during learning can drive deep exploration, and that self-imitation learning and exploration methods can be complementary.
Our results suggest that there can be a certain learning stage where exploitation is more important than exploration or vice versa. 
Thus, we believe that developing methods for balancing between exploration and exploitation in terms of collecting and learning from experiences is an important future research direction. 

\clearpage
\section*{Acknowledgement} 
This work was supported by NSF grant IIS-1526059. Any opinions, findings, conclusions, or recommendations expressed here are those of the authors and do not necessarily reflect the views of the sponsor.

\bibliography{references}
\bibliographystyle{icml2018}
\clearpage
\onecolumn
\appendix
%\clearpage
\section{Hyperparameters}

% \begin{table}[H]
% \small
% \centering
% \caption{A2C+SIL hyperparameters on Key-Door-Treasure domain. }
% \label{tab:hyper-maze}
% \begin{tabular}{l l}
% \toprule
% Hyperparameters & Value \\
% \midrule
% Architecture & FC(64) \\
% Learning rate & 0.0007 \\
% Number of environments & 16 \\
% Number of steps per iteration & 5 \\
% Entropy regularization ($\alpha$) & 0.03 \\
% \midrule
% SIL update per iteration ($M$) & 4 \\
% SIL batch size & 512 \\
% SIL loss weight & 1 \\
% SIL value loss weight ($\beta$) & 0.01 \\
% Replay buffer size & 1000 \\
% Exponent for prioritization & 0.6 \\
% Bias correction for prioritized replay & 0.4 \\
% \bottomrule
% \end{tabular}
% \end{table}

\begin{table}[H]
\small
\centering
\caption{A2C+SIL hyperparameters on Atari games.}
\label{tab:hyper-a2c}
\begin{tabular}{l l}
\toprule
Hyperparameters & Value \\
\midrule
Architecture & Conv(32-8x8-4)\\
& -Conv(64-4x4-2)\\
& -Conv(64-3x3-1) \\
& -FC(512) \\
Learning rate & 0.0007 \\
Number of environments & 16 \\
Number of steps per iteration & 5 \\
Entropy regularization ($\alpha$) & 0.01 \\
\midrule
SIL update per iteration ($M$) & 4 \\
SIL batch size & 512 \\
SIL loss weight & 1 \\
SIL value loss weight ($\beta^sil$) & 0.01 \\
Replay buffer size & $10^5$ \\
Exponent for prioritization & 0.6 \\
Bias correction for prioritized replay & 0.1 for hard exploration experiment (Section 5.3) \\
& 0.4 for overall evaluation (Section 5.4) \\
\bottomrule
\end{tabular}
\end{table}

\begin{table}[H]
\small
\centering
\caption{PPO+SIL hyperparameters on MuJoCo.}
\label{tab:hyper-ppo}
\begin{tabular}{l l}
\toprule
Hyperparameters & Value \\
\midrule
Architecture & FC(64)-FC(64) \\
Learning rate & Best chosen from \{0.0003, 0.0001, 0.00005, 0.00003\} \\
Horizon & 2048 \\
Number of epochs & 10 \\
Minibatch size & 64 \\
Discount factor ($\gamma$) & 0.99 \\
GAE parameter ($\lambda$) & 0.95 \\
Entropy regularization ($\alpha$) & 0 \\
\midrule
SIL update per batch & 10 \\
SIL batch size & 512 \\
SIL loss weight & 0.1 \\
SIL value loss weight ($\beta$) & Best chosen from \{0.01, 0.05\} \\
Replay buffer size & 50000 \\
Exponent for prioritization & Best chosen from \{0.6, 1.0\} \\
Bias correction for prioritized replay & 0.1 \\
\bottomrule
\end{tabular}
\end{table}

\section{Performance on Atari Games}
\begin{table}[H]
\caption{Performances on 49 Atari games with 30 random no-op after 50M steps of training (200M frames).}
\small
\begin{center}
\begin{tabular}{ l| r r r }
\toprule
& A2C& ACPER & A2C+SIL\\
\midrule
Alien& 1859.2& 390.2& \textbf{2242.2}\\
Amidar& 739.9& 424.8& \textbf{1362.0}\\
Assault& \textbf{1981.4}& 818.2& 1812.0\\
Asterix& 16083.3& 3533.1& \textbf{17984.2}\\
Asteroids& 2056.0& 1780.1& \textbf{2259.4}\\
Atlantis& 3032444.2& 58012.5& \textbf{3084781.7}\\
BankHeist& \textbf{1333.7}& 1203.2& 1137.8\\
BattleZone& 10683.3& 15025.0& \textbf{25075.0}\\
BeamRider& \textbf{3931.7}& 2602.4& 2366.2\\
Bowling& 31.2& \textbf{59.3}& 31.1\\
Boxing& 99.7& \textbf{100.0}& 99.6\\
Breakout& \textbf{501.6}& 118.5& 452.0\\
Centipede& 3857.8& \textbf{7790.1}& 7559.5\\
ChopperCommand& 3464.2& 1307.5& \textbf{6710.0}\\
CrazyClimber& 129715.8& 19918.8& \textbf{130185.8}\\
DemonAttack& \textbf{18331.4}& 4777.5& 10140.5\\
DoubleDunk& -0.5& -9.8& \textbf{21.5}\\
Enduro& 0.0& \textbf{3113.3}& 1205.1\\
FishingDerby& 39.1& \textbf{59.8}& 55.8\\
Freeway& 0.0& 31.4& \textbf{32.2}\\
Frostbite& 339.5& 2342.5& \textbf{6289.8}\\
Gopher& 9358.5& 3919.5& \textbf{23304.2}\\
Gravitar& 329.2& 627.5& \textbf{1874.2}\\
Hero& 28008.1& 13299.1& \textbf{33156.7}\\
IceHockey& -4.3& \textbf{0.0}& -2.4\\
Jamesbond& 399.2& \textbf{598.1}& 310.8\\
Kangaroo& 1563.3& \textbf{5875.0}& 2888.3\\
Krull& 8883.9& \textbf{11323.2}& 10614.6\\
KungFuMaster& 32507.5& 20485.0& \textbf{34449.2}\\
MontezumaRevenge& 5.8& 0.0& \textbf{1100.0}\\
MsPacman& 2843.4& 1016.0& \textbf{4025.1}\\
NameThisGame& 11174.2& 2888.0& \textbf{14958.2}\\
Pong& 20.8& 20.9& \textbf{20.9}\\
PrivateEye& 210.8& 100.0& \textbf{661.2}\\
Qbert& 17605.2& 657.2& \textbf{104975.6}\\
Riverraid& 13036.0& 2224.5& \textbf{14306.1}\\
RoadRunner& 39874.2& 8925.0& \textbf{57071.7}\\
Robotank& 3.2& 7.7& \textbf{10.5}\\
Seaquest& 1795.2& 804.5& \textbf{2456.5}\\
SpaceInvaders& 2466.1& 729.5& \textbf{2951.7}\\
StarGunner& \textbf{57371.7}& 1107.5& 31309.2\\
Tennis& \textbf{-10.3}& -17.0& -17.3\\
TimePilot& 5346.7& 3952.5& \textbf{10811.7}\\
Tutankham& 305.6& 270.7& \textbf{340.5}\\
UpNDown& 48131.8& 9562.5& \textbf{53314.6}\\
Venture& \textbf{0.0}& \textbf{0.0}& \textbf{0.0}\\
VideoPinball& 391241.6& 21797.7& \textbf{461522.4}\\
WizardOfWor& 4196.7& 1550.0& \textbf{7088.3}\\
Zaxxon& 124.2& 4278.8& \textbf{9164.2}\\
\bottomrule
\end{tabular}
\end{center}
\end{table}

\clearpage
\begin{figure}
	\centering
	\includegraphics[width=0.95\linewidth]{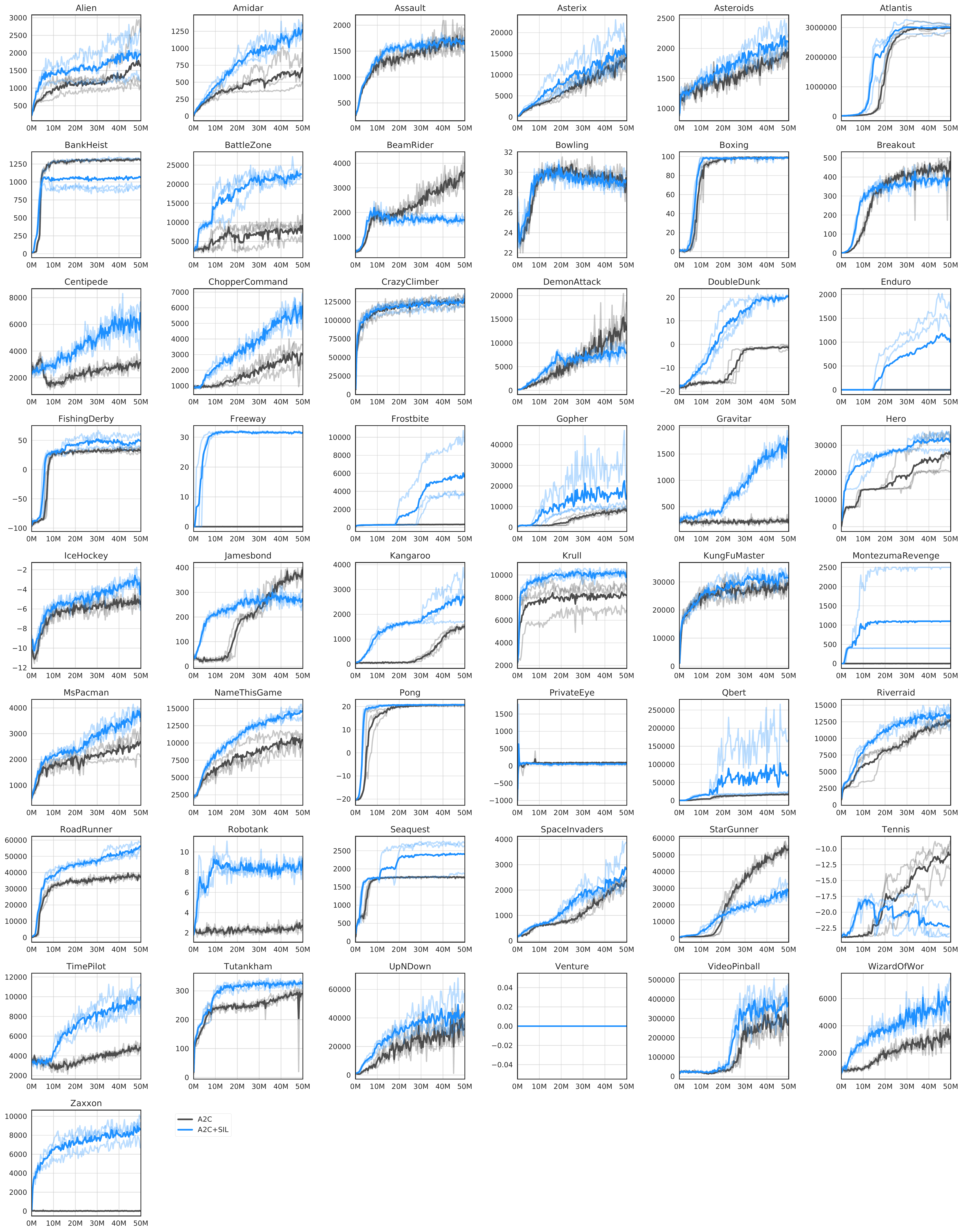}
	\vskip -0.1in
	\caption{Learning curves on 49 Atari games. }
	\label{fig:full-curve}
\end{figure}
\end{document}